# Deep-Learning-based Counting Methods, Datasets, and Applications in Agriculture – A Review

Guy Farjon[1]  Liu Huijun[2]  Yael Edan[1]

**Abstract** The number of objects is considered an important factor in a variety of tasks in the agricultural domain. Automated counting can improve farmers' decisions regarding yield estimation, stress detection, disease prevention, and more. In recent years, deep learning has been increasingly applied to many agriculture-related applications, complementing conventional computer-vision algorithms for counting agricultural objects. This article reviews progress in the past decade and the state of the art for counting methods in agriculture, focusing on deep-learning methods. It presents an overview of counting algorithms, metrics, platforms and sensors, a list of all publicly available datasets, and an in-depth discussion of various deep-learning methods used for counting. Finally, it discusses open challenges in object counting using deep learning and gives a glimpse into new directions and future perspectives for counting research. The review reveals a major leap forward in object counting in agriculture in the past decade, led by the penetration of deep learning methods into counting platforms.

✉ guyfar@post.bgu.ac.il

[1] Department of Industrial Engineering and Management, Ben-Gurion University of the Negev, Beer Sheva, Israel

[2] College of Metrological Technology and Engineering, China Jiliang University, Hangzhou, China

# Introduction

The rapidly increasing global population requires significant intensification of food production, while properly utilizing available resources, to maintain high nutritional quality and production efficiency (Kamilaris, 2018). The way forward lies in precision agriculture – the use of advanced technologies that include sensors, such as RGB-Depth (RGB-D) cameras and multi-spectral and hyper-spectral sensors; platforms, such as robots and Unmanned Aerial Vehicles (UAVs); advanced information systems, such as the Internet of Things (IOT), and Geographic Information Systems (GIS); algorithms; and work methodologies, such as data-driven management software, big-data analysis, and fusions of multiple-information sources. Precision agriculture also involves optimizing available resources to increase productivity, profitability, quality, and sustainability, to reduce the negative impacts of agriculture's footprint, and to provide nutritional security (Gebbers, 2010). In precision agriculture, valuable information is provided by counting which can provide the temporal, spatial and individual data; for example, counting the number of plant leaves, fruits, and flowers, or even plants or trees per se, can aid farmers to systematically monitor growth and predict yield and plant health and hence to make better, data-driven decisions regarding localized agricultural treatments. This leads to improved product quality and quantity by better cultivation and management practices, enabling optimal harvesting timing decisions and fast and localized disease and stress detection. In plant breeding, counting objects such as leaves, fruits, or flowers can provide a valuable tool for automatically assessing a specific cross or cultivar. In precision livestock management, counting enables farmers to obtain accurate information regarding animal living conditions that could be applied to improve animal health and welfare, leading to improved quality, production, and profitability (Ruiz-Garcia, 2009).

Traditionally, counting is done manually, in some cases by agricultural experts. This process is prone to error and is also slow and tedious, taking significant time to yield sufficient data, with all these factors combining to yield low-precision results and high labor costs (Maheswari, 2021). Moreover, when an application requires experts to be present in the field, these experts may struggle to fulfill the required demand for accurate information in a short time window. For example, the process of chemically thinning apple orchards requires the presence of experts, who have only a few days (the flowering period) to estimate the flowering intensity (Farjon, 2019). Hence, manual counting may constitute an obstacle preventing farmers from reaching informed, data-driven decisions.

Dependence on manual counting can be reduced through the use of artificial intelligence (AI)-based systems. Such systems include the use of autonomous platforms equipped with sensors and state-of-the-art machine learning (ML) algorithms to automatically collect data. ML is a sub-field of AI, in which a well-labeled dataset is used to train a set of algorithms for a specific task. ML algorithms are trained to find patterns in the trained dataset. To solve a task, the algorithm requires a labeled dataset, a cost function (from which the algorithm receives feedback on its success/failure) and a measurement metric. Once a trained model is obtained, it can be used on real-world data, and thereby reduce the need for manual counting.

When developing a counting component for a given application, the common data input is imagery data acquired using different sensors, such as RGB or RGB-D cameras, IR sensors, or multi- or hyper-spectral sensors. Due to the different types of data from these sensors, each requires the development of different, data-specific algorithms. There are two main approaches to addressing this task. The first approach is to extract features from the image. These features must be defined and can include the texture, shape, color, and size of the target object, as well as image processing statistics, such as color-based features or morphological features (Harel, 2020b). The objects are then detected from these features, followed by the counting operation. Usually, traditional ML algorithms are employed, such as k-means, Random Forest (RF), and Support Vector Machine (SVM) algorithms, and there is an abundance of papers describing the use of these methods to automate the counting procedure in agricultural applications (Albuquerque, 2019; Alharbi, 2018; Bao, 2022; Gutiérrez, 2019; Kim, 2018; Syazwani, 2022; Zhang, 2020). For example, automatic counting of fingerlings was achieved by applying a computer vision pipeline that included binarization of the image using a Mixture of Gaussian (MoG) algorithm (Zivkovic, 2006) and a blob detection algorithm that identifies geometric features based on pre-tuned parameters (such as illumination, and the shape of the fish) (Albuquerque, 2019). In a different example, a methodology for the identification and counting of pineapple crowns from top-view images captured with a UAV equipped with an RGB camera was based on algorithms that included a feature extraction stage using the RGB information as well as on geometric characteristics, such as shape and texture (Syazwani, 2022). The extracted features were put through a feature selection filter, using one-way Analysis of Variance (ANOVA), and after the features had been selected, different ML algorithms [e.g., Artificial Neural Network (ANN), RF, SVM, XGboost, LightGBM, Generalized Mixture Model (GMM), and others] were used to classify each object as a pineapple or as a false positive example. The final count estimator was the number of detected objects, as determined in the classification stage of the pipeline. In yet another example, apple clusters were classified into 7 classes by binarizing the image based on color information and feeding the output binary feature image into a classification model, based on AlexNet (Krizhevsky, 2012), under the assumption that a cluster has 0-6 apples (Häni, 2018). Finally, counting of pomegranate fruits on 3D images of pomegranate trees was achieved using morphological methods and a clustering algorithm (Zhang, 2021).

In all the above studies, the main problem associated with the ML algorithms was poor generalization, (Yamamoto, 2014; Farjon, 2019; Linker, 2017), i.e., the lack of ability to use the same settings of the proposed algorithms on other datasets. Dataset variability is inherent in agriculture (Kurster, 2018) and may be attributed to several factors, particularly: completely different object types (fruit and livestock), different versions of the same object (green and red apples; different pepper cultivars that differ in shape, color, and size), illumination variability, different sensors, an orchard vs. a controlled environment, and even changes occurring under identical growing conditions along the course of a particular year. These challenges are discussed in detail as part of this review, along with solutions to tackle them.

The second approach is to apply the algorithm directly on the image. This approach is based on deep learning (DL) algorithms, where DL is a sub-field of ML. DL is a learning approach that uses hierarchically structured multiple layers to extract useful representations. The vast majority of work in DL uses neural networks as the computational component, and when imagery data is involved, these are usually Convolutional Neural Networks (CNNs). In the past five years, ML algorithms have advanced dramatically, leading to an abundance of work in the precision agriculture domain, which, in turn, has experienced a major leap forward, mainly due to improved and well-deployed DL algorithms and computer-vision techniques. The agricultural applications in which counting using DL is widely used are crop monitoring and growth analysis, crop disease detection, livestock monitoring (Tian, 2019; Xu, 2020), insect detection (Zhong, 2018), flower density estimation for chemical thinning (Farjon, 2019), yield estimation (Linker, 2017; Xiong, 2019), and others.

In this article, we present a critical review of the counting research conducted in the past decade with focus on different objects, applications, sensors, platforms, algorithms and environments, and their setting variations. We explore different methods for counting objects in agricultural applications and present the state-of-the-art methods that exemplify the deployment of DL counters in different applications and address future perspectives for DL-based counting methods in agriculture.

# Literature Review: Methodology and Results

## Methodology

This review of scientific publications on counting methods, datasets, and applications in agriculture was performed using a five-step bibliometric analysis based on (Brereton, 2007). The five steps included: (a) selecting the databases to be searched; (b) defining the search criteria (keywords, time frame etc.); (c) first coarse screening of the papers; (d) selecting the final papers for thorough reading; and (e) analyzing and discussing the results. The process is described below in further detail.

**Selection of databases -** This review was carried out using two databases - **Web of Science (WoS)** (which includes IEEE Xplore, ACM digital library, MDPI, Springer, Elsevier, SAGE, John Wiley, Taylor and Francis, Science Direct and Google Scholar), and **Scopus**.

**Search criteria -** The search was open to any paper in the agricultural domain, but specifically those papers with a counting component. The search terms "counting" and "agri*" were used, where the asterisk was used as a means to retrieve all possible derivatives of the word "agriculture", such as "agricultural",

"agriculturist", and so on. The search was limited to peer-reviewed papers from January 1, 2012 to June 30, 2022, and was confined to papers written in the English language[1].

**Coarse screening -** The initial search identified in 2290 papers, 162 in Scopus and 2128 in WOS, including articles, proceedings papers and reviews. The initial screening included a careful reading of the title and abstract of the paper to ensure its relevance. This process narrowed the number of papers to 282.

**Final selection -** To ensure that each paper was relevant to the topic, i.e., to ensure that the paper included the use of ML algorithms to solve a counting problem in agriculture, we read the complete papers with care. This process led to the final selection of 243 peer-reviewed papers, consisting of 22 reviews and 221 articles.

**Analysis and discussion -** To follow the evolution of methods for the automation of the counting task in agriculture, we present the number of papers with respect to the type of objects, sensors, algorithmic approaches, application areas, platforms, environments, and setting variations, as published over the years.

## Literature review results

The results are divided into review papers and application papers. The results reveal that over the years there has been an increase in the number of papers that use a counting component in agriculture (Fig. 1).

**Results for the review papers –** The literature search revealed a total of 22 review papers that addressed the following topics: fruit yield estimation, insect pest monitoring, animal and crop monitoring, and phenotyping. Only five of these reviews adopted a systematic bibliographical method. The review papers, which have doubled in the past year (8 of the reviews were conducted in 2021), focused mostly on ML in agriculture in general, on the incorporation of computer vision and image processing algorithms in agriculture, and on the use of DL for various tasks. None of the 22 reviews ***targeted*** the issue of counting objects in the agricultural domain, but they did include information on the following topics:

- Progress in and the limitations of using UASs for cattle monitoring (Barbedo, 2018);

- ML-based methods for detecting and counting animals using remote sensing images (Hollings, 2018);

- Applications of fruit counting (Mavridou, 2019), but without any generalization of the counting methods;

- Use of ML algorithms to predict crop yield, with special emphasis on palm oil yield prediction, including a section on counting oil palm trees (Rashid, 2021);

- DL models for fruit detection and localization in fruit counting (Koirala, 2019);

- Applications of DL for analysis of dense scenes in agriculture (Zhang, 2020);

- Fruit yield estimation based on semantic segmentation by DL (Maheswari, 2021);

- Direct and indirect methods for fruit yield prediction and estimation in orchards (He, 2022).

The above papers, as well as papers (Anderson, 2021; Chen, 2019; Darwin, 2021; Dhaka, 2021; Hassler, 2019; Jayasinghe, 2021; Oghaz, 2019; Palacios, 2020; Saleem, 2021; Santos, 2019), mentioned the importance of object counting using ML and/or DL in agriculture. However, none of the papers provides an overview or an analysis of the counting methods. The main insight that may be drawn from these papers is that ML is currently the most widely used method, and special attention should be given to resolving counting errors. Another important conclusion that may be reached from reviewing these papers is that DL alone cannot provide accurate counting results due to the invisibility and occlusion of the targets in a cluttered environment, and therefore DL should be combined with other methods.

---

[1] Results are presented for the past decade; since the analysis is based on the number of papers published up to June 2022, the data for the year 2022 is extrapolated and will be updated as additional papers are accepted for publication.

**Results for the application papers** – The following measures were analyzed in the 221 selected papers: the main application areas, the annual increase in the numbers of papers, the targeted object/s, data acquisition techniques, platforms, and sensors, and methods and algorithms employed, as discussed below and presented in Fig. 2.

- **Targeted objects -** The number of papers containing a counting component has risen steeply in the past 5–6 years, with the early papers addressing counting methods that dealt mainly with five different types of object: fruits, flowers, trees, insects, and cereals. The categories and types of research objects are also increasing; to date, we have observed a total of 9 main target object categories, namely, livestock, leaves, vegetables, crop plants, and the aforementioned five applications (Fig. 2A).

- **Sensors** - Different sensors are used for the acquisition of image information to count objects in precision agriculture (Fig. 2B). These include Light Detection and Ranging (LiDAR) sensors and different types of cameras: monochrome, color, RGB (including mobile phone cameras), RGB-D, multi-spectral/hyper-spectral, and thermal cameras. In several studies, a combination of multiple sensors was used to better analyze the traits of the targeted object. For more than a decade, RGB cameras were used in most papers (82.93%), probably mainly due to their low price and the increase in the application of mobile phone cameras. Nonetheless, there has been an increase in attention to this field, with a small, but growing, number of papers using other types of sensors. This increased interest indicates that there is a solid basis for future research in this direction.

- **Algorithmic approach -** Prior to the incorporation of ML and DL in the agricultural domain, object counting was performed mainly by extracting basic features from an image using morphological operations, such as size, shape, color, texture, etc. Currently, there are three main algorithmic approaches to counting objects in images: i) conventional image processing methods based on morphological operations; ii) ML algorithms using features extracted from the image; and iii) DL methods applied directly to the input images. As can be seen in Fig. 2C, DL-based methods have dominated the field in the past few years, probably due to their relatively easy implementation and excellent performance results.

- **Application areas -** The count-related application areas of the papers may be classified into five categories (Fig. 2D): yield estimation (51%), phenotyping (28%), livestock monitoring (11%), insect monitoring (7%) and others (3%). The category 'others' includes applications related to the field management of orchards, such as pruning (Santoro, 2013; Westling, 2021), path planning (Mokrane, 2019), plant water status (Bruscolini, 2021), monitoring fruit ripeness (Tenorio, 2021) and monitoring resources (Vermote, 2020; AlMaazmi, 2018).

- **Platforms -** We also examined the use of different platforms in the papers across the years (Fig. 2E). The dominant approach to collecting data is manual (i.e. using a tripod or manually capturing images), but the use of automatic data collection (using robots or UAVs) has increased in recent years, enabling massive data collection. Automatic data collection introduces additional challenges, as we discuss below in the Challenges section.

- **Settings variations -** Research has gradually developed from a single variety, time, and place to multi-variety, multi-year and multi-production areas, including crops in different growth stages (Fig. 2F). This important evolution in the field makes the recent methods more robust and generalized for different conditions. The importance of developing methodologies and algorithms able to address variations in settings is discussed further in the Challenges section.

- **Environments -** Applications such as fruit counting, grain yield estimation, and animal monitoring are carried out mainly in outdoor environments, basically in a field or in a completely open environment in the wild, i.e., under natural conditions. Only a few studies were carried out in indoor environments or laboratories, and all focused-on crop phenotyping.

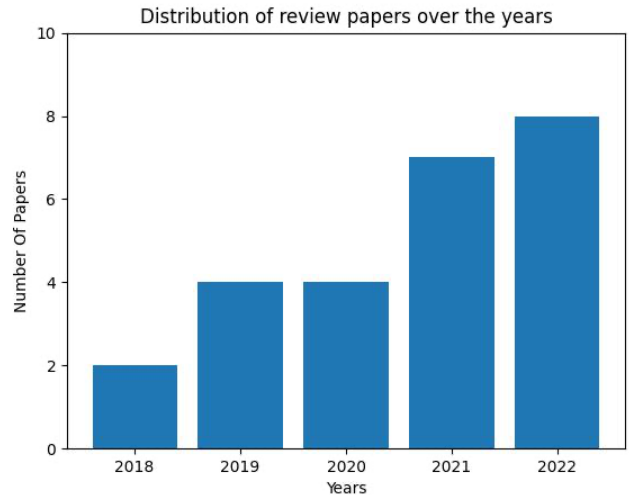

**Fig. 1** Distribution of review papers dealing with counting in agriculture based on deep learning. Data in 2022 was extrapolated based on first six months. There were no review papers between 2012 and 2017.

## Summary of the Literature Review Results

The results of our literature search revealed a rapid increase in the attention being paid to counting objects in agricultural environments, with the number of papers including a counting component rising steeply. This rise is probably due to the 'maturity' of data acquisition technologies (sensors and platforms) and the advantages provided by ML architectures and frameworks. From a general perspective, it may be said that when data collection is easy (using UAVs or robots), then the number of papers that include a DL component also increases. From a sensor perspective, the literature search showed that there are only a few papers that used advanced sensors (such as hyper-spectral/depth cameras) for counting objects. We thus see great opportunities for researchers and practitioners in developing DL architectures using advanced sensors.

## Deep-Learning Counting Methods

In this section, different DL counting methods are introduced, including full explanations together with the data collection and annotation burden. For each method, several relevant papers are reviewed.

### Direct Counting

There are two approaches for using CNNs to directly count the number of objects in an image. When using a classifier, some domain knowledge is needed to bound the number of classes for classification. For example, Häni et al. (2020) used a Resnet50 classifier to count apples from image patches containing apple clusters. Since apple clusters contain up to six apples, the authors defined the problem as a 7-classes classification problem, including zero apples in a patch. In that paper, a segmentor was first applied to extract Regions of Interest (ROIs), and then the number of apples in the ROI was predicted using the 7-classes classifier.

A more robust method is to use a deep regressor, in which the number of classes is not bound. A deep regressor is a deep CNN network (mostly previously designed for classification problems), with a regression head (instead of the regular classification head). In this method, a set of images and a number of objects for each image, $[I_j, y_j]^N_{j=1}$, is introduced. The deep regressor has only the number of objects as the supervisor, which simplifies the collection of annotated datasets. However, it is unclear what the model learns to recognize, which is probably the reason that this deep network method is less commonly employed.

The ResNet50 backbone (He, 2016) was also used to count leaves for the Leaf Counting Challenge (LCC) publicly available dataset (Dobrescu, 2017), with the final outcome of the 2017 LCC being a state-of-the-art leaf counter, developed as follows. The Challenge showed that a simple direct regressor trained with a limited number of images is able to learn how to count. The direct regressor achieved 0.91 in the reported metric, namely, the absolute Difference in Count ($|DiC|$), which is similar to the Mean Absolute Error (MAE). Based on this strategy, a more complex pipeline using a direct regressor was developed (Farjon, 2021). Previously, a Multiple Scale Regression (MSR) pipeline had been built on the RetinaNet (Lin, 2017) architecture, but with an important addition to the loss function. The uncertainty loss function from Kendall & Gal (2017) was applied to learn how to count leaves on multiple scales [based on the Feature Pyramid Network (FPN) output)]. The results were fused using a Maximum Likelihood Estimation (MLE) function. Results on the same LCC dataset achieved 0.83 in the $|DiC|$ score, thereby introducing a state-of-the-art method for counting leaves.

Due to the difficulty in comprehensively understanding the results of the above models, Dobrescu (2019) set out to analyze the results obtained with these deep regressors. For this task, they used the VGG-16 network (Simonyan, 2014) as their regressor and visualized areas in which the images were relevant for the final count by using Layerwise Relevance Propagation (LRP) (Bach, 2015) and Guided Back Propagation (GBP) (Springenberg, 2014). Their methodology was based on introducing into the network images from the training set on which they had placed differently sized "black-squares" to simulate occlusions of leaves. Their analysis showed that the regressor's middle layers focused on the plants' edges, while in the top (fully connected) layers, the most active neurons focused on different areas in the image. The results thus showed that the trained model does not store information about the leaf number in an image, but rather focuses on leaf edges when counting leaves.

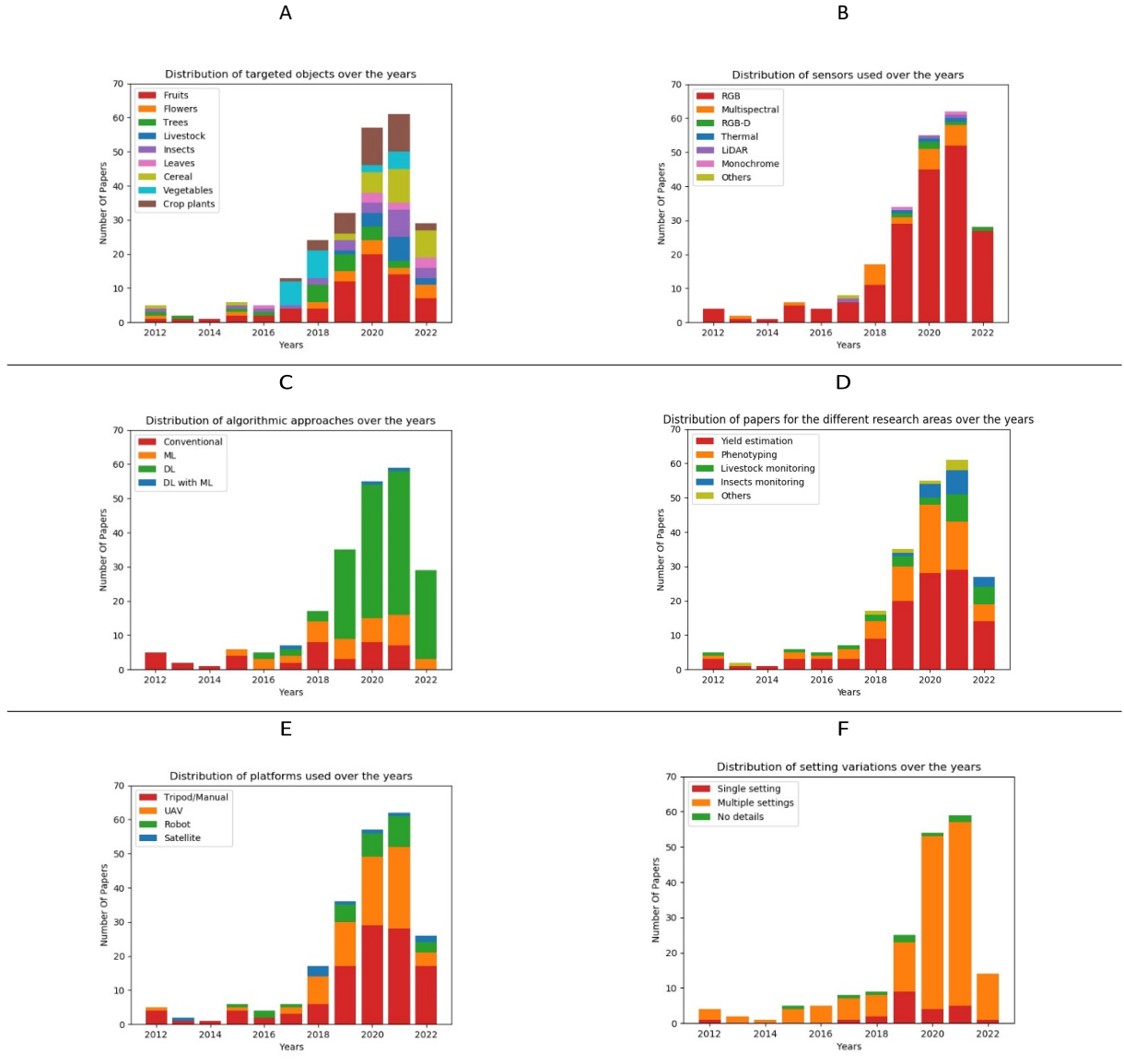

**Fig. 2.** Literature review results. Data for 2022 includes only the first six months. A - Distribution of targeted objects over the years, where the two main object categories are fruits and vegetables. B - Distribution of sensors used over the years. C - Distribution of algorithmic approaches over the years. D - Distribution of papers for the different research areas over the years. E - Distribution of platforms used over the years. F - Distribution of setting variations, for single variation vs. multi-variation, over the years.

The use of a weakly supervised approach for counting apple fruits and flowers, using only image-level count annotation, was recently demonstrated (Bhattarai, 2022). That study proposed a regression-based network, known as *CountNet*, for flower and fruit counting in outdoor environments. The network learnt from minimal feature information on the object of interest without explicit detection and localization. The image-active regions and features were visualized with the aim to demonstrate the interpretability and explainability tools.

### Detection-based and segmentation-based counters

Object detector-based counters incorporate an object detector as the counting component. This approach is natural, since the algorithm is designed to detect target objects in the image. These detected objects are then counted, and their number is the final estimated count. The object detection task consists of determining the location in the image where certain objects are present and classifying them into specific target classes. At the training stage, the object detector receives the set $[I_j, B_j]^N_{j=1}$, where $I$ denotes a set of images, and $B_j$ denotes a set of bounding boxes surrounding each object in each image $i$. The size of $B$ can vary from image to image. Each $[b_k \in B_j]^{|B_j|}_{k=1}$ holds four parameters describing the box dimensions and one additional number stating the object's class. At test time the model outputs a set of detection hypotheses $D$, in which each of the proposed bounding boxes contains the target object, and a confidence score that the bounding box contains an object (for multi-class detection, the network will also predict the specific class); see the Metrics sections for additional explanations. The annotation burden in object detection-based methods is relatively high, since the annotation itself is more difficult (relatively to the object count in regression-based counters), but even more so for addressing the challenges introduced in the agricultural domain. As can be seen in Fig. 3, the images include variations in the objects' size and color, and many of the objects are occluded or blurry. The different variations are discussed in detail in Challenges section.

Segmentation-based counters incorporate an object segmentation module as the counting component. In this approach, the objects are not only detected but also segmented before counting. To reduce false alarms and improve the counters' results, the segmented objects can be further processed before counting. The object segmentation task consists of determining the location in the image where certain objects are present, classifying them into the specific target classes, and segmenting the specific area of each object. DL-based object detectors have produced excellent results in recent years on various public datasets (not specifically related to agriculture), such as Pascal-VOC (Everingham, 2010), MS-COCO (Lin, 2014), and more. Due to this success and their availability in open-source repositories, object detectors have become the most common counting component for object counting in agriculture. This component is mostly used as is, without any improvements or changes, as in the examples that follow.

A comparative study of the Faster-RCNN (Ren 2015), SSD (Liu, 2016), MobileNet (Howard, 2017), RetinaNet (Lin, 2017), and Efficient-Det (Tan, 2020) ($D_0$ and $D_4$) object detectors was conducted for the task of detecting and counting *Matsucoccus thunbergianae* insects caught in insect traps (Hong, 2021). The size of the insects is relatively small (60×60 bounding boxes), compared to the image resolution (6000×4000), and such small objects are difficult to detect, since object detectors often resize images to smaller resolutions (up to 1024×1024 in the above networks). This challenge is usually solved by cropping the image into $K$ sub-images, as was done by Hong et al. (2021). This solution was also used in (Farjon, 2019), in which the author used Faster-RCNN (Ren, 2015) first to detect apple flowers, and then to count them, to estimate the flowering intensity during the flowering period, and finally to provide an estimation the peak flowering day. In Mosley et al. (2020), sorghum heads in RGB images captured by a UAV were counted, yielding good results by using a tuned anchor box for the Yolov3 (Redmon, 2018) architecture. The tuning process of the anchor was necessary due to the difference in object size between the COCO dataset and the sorghum head datasets, and specifically, a 10×10 anchor was added, to capture small sorghum heads.

A Mask-RCNN model was trained to segment blueberries in a cluster for yield prediction (Ni, 2020). Four different blueberry cultivars ('Emerald', 'Farthing', 'Meadowlark', and 'Star') were detected and classified as mature or immature. The use of multiple cultivars is important for generalization, as we will show in the Challenges section. To measure the model's ability to correctly detect the blueberries, a linear regression model was fitted between the manual counting and the predicted number of berries in a cluster.

A two-step pipeline was used to detect and count wheat spikes (Sadeghi-Tehran, 2019). The process was started by extracting wheat spike candidates using Simple Linear Iterative Clustering (SLIC) to reduce the complexity, i.e., the number of pixels, fed into the CNN. The second step was conducted to differentiate between different clusters, namely, to count the number of wheat spikes by using a CNN architecture. A two-step algorithm was used in (Palacios, 2020) to quantify the number of grapevine flowers and to find a correlation between the estimated number of flowers and the actual yield per vine. Data was collected on-the-go under field conditions by using a mobile sensing platform. First, a mask of grapevine flowers was extracted using two SegNet semantic segmentation frameworks. Then, based on the segmented masks, the number of flowers was estimated using a linear regression model.

For a better definition of sorghum panicle boundaries, Malambo et al. (2019) developed a field-based approach for counting panicles from UAS images by using SegNet for supervised semantic segmentation.

In post-processing, morphological operations were used to remove small objects and separate any merged panicles for individual counting. The edges were poor due to the down-sampling in the encoder of SegNet, (which is also present in the U-Net network). The resolution of the images and the mosaic of the UAV images also had impact on the segmentation, but the impact was less marked for the counting problem.

To achieve a higher level of automation, some researchers acquire video data from the fields. The idea is that instead of a person using a camera, the data could be collected using a UAV. For example, videos of vineyards to detect, segment, and track grapes clusters have been used (Santos, 2020). Mask R-CNN was applied for individual grape cluster detection and Structure-from-Motion (SfM) for instance matching and tracking. In a different study, RetinaNet (Lin, 2017) was used to count pistachios in videos (Rahimzadeh, 2022). In that study, the camera was static and the objects were moving on a transportation line, enabling analysis from multiple images. When new objects entered the frame, the algorithm matched "old" pistachios in consecutive images and also found new pistachios. The main task was to count open and closed pistachios, and therefore an additional classifier function was applied. This additional step was needed because the pistachios were turning and spinning on the transportation line, causing the detection algorithm to classify the same objects to different classes in different frames. The methodology enabled tracking of the detected objects and hence counting of the total numbers of open and closed pistachios. In yet another study (Anderson, 2021), fruit load was estimated for fruit collected from 37 orchards in 2019-2020 and from 19 orchards in 2020-2021. Data was collected for trees of different ages and cultivars and for different orchard layouts. Fruits were detected across each frame of the collected videos using the MangoYolo detector (Koirala, 2019). To track the fruits between frames, the Kalman-Filter method was adopted.

The above methods all used RGB images as input data. However, they were not always adequate for dealing with specific challenges; in such cases, researchers tried to improve the results by introducing additional information to the object detector/segmentor. For instance, RGB-D cameras or multi-spectral sensors are used to add information to the learning process or to extract additional features from images. Since the multi-spectral camera captures additional information, it can be used for crop maturity evaluation and refined segmentation, disease detection, and estimation of growth status. For example, a method for counting cherry tomatoes at different maturity levels was developed using Fast R-CNN for detection on images acquired by a RGB camera, followed application of the DeepSort algorithm for multi-target tracking to prevent double counting (Chen, 2020). By alignment of the RGB and the NIR reflectance (acquired by the multi-spectral camera) images, a normalized difference vegetation index (NDVI), as a metric of maturity, was obtained for each tomato by using an SVM classifier, thereby enabling the counting of tomatoes by maturity level. In a different example, DL was used to estimate the number of citrus trees in highly dense orchards from UAV multi-spectral images (Osco, 2020). In such high-density plantations, where the boundaries of individual plants are not sufficiently distinct to facilitate separation between trees, the detection performance degrades as the plants decrease in size. A dense confidence map of the probability that a plant exists in each pixel was developed to estimate the number and geolocation of the citrus trees, rather than the object-detection method using rectangles to represent trees. Afonso et al. (2020) used DL instance detection and segmentation of tomatoes in a greenhouse at night by using Mask R-CNN. They used the depth channel for post-processing of the results, discarding background false positives. This procedure improved the precision, but reduced recall, possibly due to the low precision of depth values from depth images, causing some of the foreground fruits to be excluded.

An end-to-end procedure was developed to detect and count passion fruits in RGB-D images (Tu, 2020). The images were captured in a natural, outdoor environment, which is important for generalization. Since passion fruits are relatively small, a Multiple Scale Faster Region-based Convolutional Neural Networks (MS-FRCNN) method was proposed, which integrates feature maps from lower layers with ROI pooling feature maps. The feature extraction network of Faster R-CNN was replaced with ResNet, combined with top-down connection. This process produced multi-scale feature maps of three scales (corresponding to conv3, conv4, and conv5 convolutional blocks, respectively) trained with RGB and depth images acquired by an RGB-D camera. Compared with the faster R-CNN detector of RGB-D images, the recall, the precision and the F1-score [2] of the MS-FRCNN method increased from 0.922 to 0.962, 0.850 to 0.931 and 0.885 to 0.946, respectively, which is a significant improvement for such small and occluded fruits. This fused-channel network gave an F1-score of 0.946, outperforming only-RGB and only-depth detectors by up to 10%.

---

[2] Recall, precision and F1-score are defined in the Metrics section.

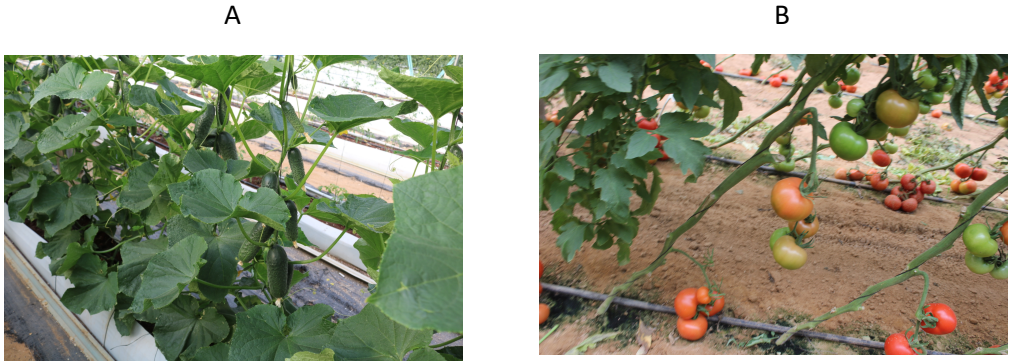

**Fig. 3** Challenges in object detection in the agricultural domain. (A) Cucumber fruits in a greenhouse; objects are of different sizes with many occlusions. (B) Tomato fruits in greenhouse; objects are of different sizes and color and objects in the second and third rows are small and blurry.

## Density estimation-based counting

Object counting using a density estimation component is considered a regression task, in which the network attempts to estimate the number of objects by predicting a heat-map representation of the image and then to regress the number of objects from it. During training, the network is fed with a set $[I_j, P_j]_{j=1}^N$, where $I$ denotes a set of images, and $P$ denotes a set of points, manually annotated, representing the center of the target object, meaning that each $[p_k \in P_j]_{k=1}^{P_j}$ holds two parameters representing a pixel in the image. The network is guided using a pre-defined heat-map, crated by the points in $P$, (e.g. in Farjon et al. (2019) placed a two-dimensional Gaussian on each center point, and the network was trained to predict these Gaussian heat-map). The annotation burden in density estimation-based counting is relatively low, since the annotator is simply required to place a single point at the center of the object. However, it is necessary to place the center-point annotation on the object's center of mass, which is not always simple to estimate, especially when objects are partially occluded or are not round.

The above-described method is not as common as detection or segmentation-based counters, based on our the reviewed papers, but is intuitive and simple to use. Since the task of object counting does not require the model to distinguish between different instances, regressing the number of objects based on a density estimation yields state-of-the-art results, as exemplified in the examples given below.

Tian et al. (2019) argued that due to the different degrees of perspective distortion, different postures, and occlusion, directly regressing the number of objects (in this case pigs) from the detection results was not accurate enough, and that the density map contained richer spatial distribution information. Therefore, a modified counting CNN model was used to learn mapping from image features to density maps, which were integrated to obtain the total number of pigs in an entire image in a real pig farm.

Another example is that of counting chickens (Cao, 2021) by using a localization-based counting algorithm, i.e., a point-supervised method, known as LC-DenseFCN. Due to the high density of chickens in poultry farms, object detection and segmentation are not suitable for chicken counting in images. Unlike other counting methods based on object detection, the LC-DenseFCN method was more efficient for counting chickens, since there was no requirement to define the full object features, especially when there were multiple chickens overlapping. In comparison with counting methods based on semantic segmentation or instance segmentation, the proposed method had a simpler architecture and reduced the annotation expenditure.

Farjon et. al. (2021) presented the Detection-Regression Network (DRN) architecture; this architecture was also built on the RetinaNet (Lin, 2017) architecture (similarly to the MSR mentioned above in the Direct

Regression section), but with three differences: i) instead of using all the pyramid levels, only the largest resolution ($C_3$ output) was used; ii) the regression sub-network was replaced with a detection based sub-network (to detect the center point annotations); and iii) the loss function was replaced with a two-term loss function, one term designated for the density estimation, and the other for regressing the final estimator. The authors were able to obtain state-or-the-art results on the LCC dataset, outperforming other image-level supervised networks.

Dijkstra et al. (2018) proposed a Fully Convolutional Neural Network (FCNN) based on U-Net architecture termed CentroidNet, for crop localization and counting from UAV images. This network produces an output with the same height and width as the input, with $C+2$ output channels (where $C$ is the number of classes). The two additional output maps, represents a coordinate of a pixel which is likely to contain a centroid. These coordinates are aggregated, creating a voting map. A high value in this map, represents a higher chance of that pixel to contain a centroid. The rest of the channels holds the logit map, similar to the output of semantic segmentation, i.e., each pixel represents the probability of it belonging to class $c_i \in C$. Thereafter, the segmented images were generated as the output, which contained the centers of the plants. The results showed that false detections, especially on low-quality images produced by drones, were suppressed by using a trained segmentation mask.

Wu et al. (2019) presented a method for in situ counting of rice seedlings from RGB UAV imagery with two FCNNs. The first was trained for segmenting rice seedling areas and non-rice seedling areas in each image, while the second was used to estimate the distribution of rice seedlings and to generate a density map, similarly to other methods presented in the Density estimation-based counting section. Additional examples (Gomez, 2021; Hobbs, 2020; Rahnemoonfar, 2019; Xiong, 2019) applying different variants of density estimation networks have also been proposed.

## Application specific methods

In this section, we present studies that either incorporate several components or use a custom-made network to solve the counting task. These studies usually try to create an end-to-end solution that is based on a data collection platform, architecture, and analysis. The methods are usually very specific and may not be fully generalized. For example, Bellocchio et al. (2019) proposed the WS-COUNT network, which counts the number of objects in the image, with binary supervision alone—the object is either present or absent. The network is trained to count general objects by trying to learn the concept of 'objectness,' and to predict the number of objects based on the whether the object is present or absent. The WS-COUNT network makes use of several components: i) a supervised regressor (S-COUNT), ii) a PresentAbsent Classifier (PAC), and iii) a Multi-Branch Counting CNN (MBC-CNN). The S-COUNT is trained separately and is not tuned in the subsequent steps. The PAC network is a simple classifier that is trained with the above-mentioned binary supervision. The MBC-CNN uses multiple scales of the image to count the objects in different resolutions. The final count is estimated by fusing the information from all the components. The WS-COUNT is a novel weakly supervised strategy that was able to learn to count without requiring task-specific supervision labels. Counting results for olives, almonds and apples were comparable to those of fully supervised baselines. Although this method does not distinguish occluded fruits well, especially when the fruits are in a dense group, it provides novel ideas for new fruit counting methods.

Following (Bellocchio, 2019) and in view of the fact that learning a new and specific model for each object is expensive and impractical (following the supervised paradigm), a weakly supervised transfer learning model was proposed to count fruits which were not included in the train set (Bellocchio, 2020). The suggested model uses a Presence-Absence Classifier (PAC) similarly to the above method (Beelocchio, 2019), following by a Cycle-Generative Adversarial Network (C-GAN) to transform images from a source orchard (i.e. an annotated dataset) to a target orchard, where no labels are available. The C-GAN model is trained to translate the objects appearance from the source orchard to the target orchard, thus enabling the model to count objects from unseen classes. The trained PAC generatd the weak supervision signal for the counting block, which could then be used on the target orchard directly.

In Nellithimaru et al. (2019), a stereo camera mounted on a robot was used to count grapes in a vineyard. The architecture included three steps: first, the images were put through the MASK-RCNN network to provide a segmentation mask (and a first count estimation) for each image; second, the produced masks were used to generate a 3D reconstruction of the scene, followed by the use of a Singular Value Decomposition (SVD) algorithm to produce a second count estimation; and third, a circular shape was fitted for each of the found graphs, followed by the Simultaneous Localization and Mapping (SLAM) algorithm to produce the final count estimation. These steps improved the counting accuracy results and boosted the inference speed.

## Datasets and Evaluation Metrics

### Public datasets

One of the major challenges in the agricultural domain is the scarcity of well-annotated datasets. In Table 1, a list of the publicly available datasets, compiled from the above in-depth literature review of 243 papers, is presented; the Table shows the type of data and the methods employed (detection/segmentation), along with links to the datasets.

### Evaluation metrics

The relevant metrics used to evaluate the performance of the counters or of other relevant components used in counting pipelines are discussed below. As stated above, the counting of objects is used for various applications, but each application requires a specific metric that will maximize the outcome of the entire pipeline.

**Counting metrics** - To evaluate the counter's performance, the deviation of the predicted value ($\hat{y}$) from the ground truth number ($y$) for each example $i$, $e_i = y_i - \hat{y}_i$, is measured. Using this error, the overall performance is evaluated by aggregating the errors over all examples.

The standard and widely used metrics for evaluating overall performance are the Mean Absolute Error (MAE) and the Mean Square Error (MSE):

$$MAE = \frac{1}{N}\sum_{i=1}^{N}|e_i| \quad (1)$$

$$MSE = \frac{1}{N}\sum_{i=1}^{N} e_i^2 \quad (2)$$

In some cases, researchers use the Root Mean Absolute Error (RMAE) or the Root Mean Square Error (RMSE), which are equivalent to the MAE or MSE. Two additional commonly used metrics are the relative Root Mean Squared Error (rRMSE) and the coefficient of determination ($R^2$). The rRMSE measures the deviation of each example with respect to the number of objects, and $R^2$ measures the strength of the relationship between the model predictions and the ground truth:

$$rMSE = \sqrt{\frac{1}{N}\sum_{i=1}^{N} \left(\frac{e_i}{y_i}\right)^2} \quad (3)$$

$$R^2 = 1 - \frac{\sum_{i=1}^{N} e_i^2}{\sum_{i=1}^{N}(y_i-\bar{y})^2} \quad (4)$$

Table 1

Publicly available datasets, as collected from the 243 reviewed papers. The annotations available in each dataset along with a link to download the data are presented.

| Target object | Data type | Count | Center-point | Detection | Segmentation |
|---|---|---|---|---|---|
| Fish [1] (Albuquerque, 2019) | Videos | ✓ | - | - | - |
| Grape clusters [2] (Santos, 2020) | RGB Images | - | - | ✓ | ✓ |
| Apple flowers [3] (Bhattarai, 2022) | RGB Images | ✓ | - | - | - |
| Apples [3] (Bhattarai, 2022) | RGB Images | ✓ | - | - | - |
| Apples [4] (Gao, 2020) | RGB Images | - | - | ✓ | - |
| Apples [5] (Häni, 2020) | RGB Images | - | - | ✓ | ✓ |
| Mangoes [5] (Häni, 2020) | RGB Images | - | - | ✓ | - |
| Almonds [5] (Häni, 2020) | RGB Images | - | - | ✓ | - |
| Grape clusters [6] (Liu, 2020) | RGB Images | ✓ | - | - | - |
| Pigs [7] (Tian, 2019) | RGB Images | ✓ | ✓ | - | - |
| Cattle [8] (Soares, 2021) | RGB Images | - | - | ✓ | - |
| Oil palm [9] (Tong, 2021) | RGB Images | - | ✓ | - | - |
| Acacia [9] (Tong, 2021) | RGB Images | - | ✓ | - | - |
| Wheat [10] (David, 2021) | RGB Images | - | - | ✓ | - |
| Oil radish [11] (Krogh, 2019) | RGB Images | - | - | - | ✓ |
| Plant seedlings [12] (Jiang, 2019) | Videos + RGB Images | - | - | ✓ | - |
| Apples [13] (Gené-Mola, 2019) | RGB-DS | - | - | ✓ | - |
| Mango [14] (Kestur, 2019) | RGB Images | - | - | - | ✓ |
| Apple [15] (Gené-Mola, 2020) | RGB Images + SfM | - | - | ✓ | ✓ |
| Vegetable (various) [16] (Lac, 2022) | RGB Images | - | - | ✓ | - |

[1-16] Links to the datasets are presented in Appendix A.

**Object detection and segmentation metrics -** When using detectors and segmentors for counting, it is important to evaluate separately the performance of the detection or segmentation task and the performance of the counting task. There are two main measures for evaluating the performance of these components, namely, 1) the mean Average Precision (mAP) and 2) the F1-score. Using accuracy / error metrics are not useful for object detection tasks, since the 'negative' class (for non-object bounding boxes) is significantly larger than the 'positive' class. The mAP metric (which is the area under the recall-precision curve) is a better alternative, taking this imbalance into account:

1. **mAP** – This measure is similar for the detection and segmentation tasks. When using an object detector, the evaluation is per object, while in segmentation, the evaluation is made pixel-wise. In this section we describe only the evaluation of object detectors.

    The mAP metric measures the precision and recall for different prediction confidence thresholds $t_p$. At test time, the detector outputs a set of detection hypotheses $D$ for each image $i \in I$. A detection hypothesis $d \in D$ is considered True Positive (TP) if the Intersection over Union (IoU) between $d$ and the ground truth $GT$ is greater than some threshold $t_{IoU}$, where, usually, $t_{IoU} = 0.5$ $\left(\frac{d \cap gt}{d \cup gt} > t_{IoU}\right)$.

In the detection task, the mAP is used to evaluate the detector's performance. mAP is evaluated as the mean score of the Average Precision (AP) for each object class. The AP score is the area under the recall-precision curve, where recall is the percentage of correct predictions out of the number of positive examples [6] and precision is the percentage of positive predictions that are correct [5].

$$Precision = \frac{TP}{TP+FP} \quad (5)$$

$$Recall = \frac{TP}{TP+FN} \quad (6)$$

Where FN are the false negative examples, obtained when the detector fails to detect annotated objects. And the FP are the false positive examples, obtained when the detector falsely detects different objects in the image and classifies them as objects. The detection hypothesis can produce a false alarm for various reasons: $d$ does not contain an object is the trivial case, but a false alarm can also occur as a result of incorrect classification of an object in the bounding box or as a result of double counting of the same object using more than a single box.

The mAP is evaluated by changing the confidence threshold ($t_p$). As stated in the Detection-based and Segmentation-based Methods section, each detection hypothesis ($d$) also receives a confidence score. When this score is greater than the confidence threshold, $d$ is declared as an object. To output a specific counting measurement, one must choose a threshold that will optimize the counting results. This threshold will be naturally chosen as the one that does not introduce counting bias. Usually, this threshold is selected so as to balance between the number of FPs (falsely detected objects) and FNs (missed objects). However, this threshold is relevant in target application, which requires only the number of objects, without a need for post-processing (as in Vitzrabin et al. (2016b)). There are cases in which other thresholds should be considered (e.g., in the case of robotic harvesting, to minimize false alarms so as to minimize harvest time).

2. **F1-score** – This measure summarizes the detector's performance at a specific confidence threshold.

$$F_1 = \frac{2 \cdot (Recall \times Precision)}{Recall+Precision} \quad (7)$$

This method is also a natural choice in counting applications, since the confidence threshold chosen for counting will be the one that yields the highest F1-score.

**Density estimation** – Here, we describe the measures used for density estimation for counting. We define density estimation-based counters in terms of their supervision, namely, a single point-annotation placed at the center of the target object (Farjon, 2021; Rahnemoonfar, 2019). For this approach, there are two main metrics, 1) the Structural Similarity Measurement Index (SSIM) and 2) the Percentage of Correct Keypoints (PCK). The SSIM metric is designed to evaluate the quality of the predicted output heat-map against the ground truth heat-map (created using the provided point-annotation, as discussed in the Density estimation-based counting section). The PCK is used to evaluate the localization of the predicted point, against the provided point-annotations, with respect to the size of the detected object:

1) The SSIM is expressed as:

$$SSIM(\hat{y}, y) = \frac{(2\mu_y\mu_{\hat{y}}+c_1)(2\sigma_{y\hat{y}}+c_2)}{(\mu_y^2+\mu_{\hat{y}}^2+c_1)(\sigma_y^2+\sigma_{\hat{y}}^2+c_2)} \quad (8)$$

where $y$ is the ground truth mask, $\hat{y}$ is the predicted mask, $\mu$ is the average of the mask, $\sigma^2$ is the variance of the mask, and $c_1$ and $c_2$ are two variables used to stabilize the division (in case the denominator becomes 0).

2) For the PCK, a predicted point is considered a hit if the distance between it and a ground truth center annotation is lower than $\alpha * max(w, h)$, where $(w, h)$ are the width and height of the bounding boxes surrounding the entire target object, and $\alpha$ is hyperparameter controlling how far is a point allowed to be from the ground truth annotation, to be considered a hit ($0 < \alpha < 1$). Has $\alpha$ approaches 0, the tolerance for inaccurate predicted points position is low (Yang, 2012).

While the SSIM provides image level analysis and does not account for the localization of the target objects, the PCK metric also accounts for the localization of the predicted object centers. Note that the PCK can be computed when the predicted output map is not dense, implying that the point is specifically predicted (Farjon, 2021).

## Challenges

From the articles reviewed here, we reached an understanding of the challenges that must be addressed when developing new methods. Here, we propose ways to move forward to yield improved DL solutions. In particular, there are two types of challenge when using DL-based counters in agriculture. The first type – the domain challenge – relates to the unstructured and dynamic nature of the outdoor agricultural environment. The second type of challenge relates to hardware aspects, such as computation devices, multi-modality constraints, and the accuracy of sensors.

### Domain challenges

Many challenges related to using DL-based counters are associated with agricultural and environmental conditions. The use of deep counters under real-world outdoor conditions (field or greenhouse) gives rise to significant challenges due to the unstructured and highly variable conditions that cannot be controlled or modeled. Furthermore, the biological nature of the product further increases the variability. Some of the challenges are general computer-vision challenges that are exacerbated in the agricultural domain, and others are caused directly by the unique characteristics of the domain. The different challenges are discussed in greater detail below.

**Target object variability** - In agriculture, target objects have high intra-class variation. Inherently, the same object will exhibit large variations caused by changes along the growth stages (Harel, 2020a), different viewpoints (causing the same fruit, plant, or tree to look different depending on the viewpoint) (Hemming, 2014), and even different features of a single object (e.g., color) (Ringdahl, 2017; Kurtser, 2018); for example, apricots are green at the beginning of the growth season but their color changes to yellow-orange as they grow. Similarly, some apple cultivars (e.g., Pink Lady) have both red and yellow segments, and sweet peppers can be red on one side and green on another, since they mature in a non-homogeneous manner (Harel, 2020a). To address this variability challenge, it is important to collect data that includes all possible variabilities and to ensure that data from different groups of objects appear in the training, validation, and test sets. An additional trait of agricultural objects (especially fruits and flowers) is that they are usually small w.r.t other objects in the environment (e.g., small fruits on a large tree). In most use cases, the target objects should be counted with respect to the plants or trees from which they originate. Furthermore, most data sets include high-resolution images (e.g., 3648×5472 in (Farjon, 2019)), such that the image completely captures the tree and its surroundings, causing the target objects to be tiny (70 × 100 in (Farjon, 2019)). Usually, the input size of the CNN is an order of magnitude smaller than the high-resolution images. Thus, the first step of the algorithm is to resize the image. As a result, the annotated bounding boxes are also resized and become too small to find. The simple naive solution is to tile the image into $K$ sub-images, with a small overlap between adjacent tiles. However, in this method, tiling the original image might lead to double counting of the same objects in different tiles. Thus, it is important to reconstruct the image from the tiles and to apply a non-maxima suppression on the original image. This issue is discussed in depth in Wosner et al. (2021).

**Variations in illumination** - An additional challenge addressed in many studies is illumination variation. Since images are captured under outdoor conditions, the position of the sun and the ambient weather conditions (clouds, rain) will have a marked impact on the captured frame. In addition, the image is likely to include leaves and branches that cause shading, which will vary along the course of the day and will depend on the viewpoint. All these factors lead to marked differences in illumination conditions between images and even within the same image (Vermote, 2020). Some objects might be in shade, some might be under direct sunlight, others might be dark, and yet others might be subject to glare. To avoid miss-detection, these problems must be considered when dealing with all agriculture-related tasks, especially for counting objects. A possible direction is to attempt to sample the same target object under different illumination conditions (different times of the day, weather conditions, i.e., cloudy vs. sunny) and to include these different samplings in the training set. Such an approach has two setbacks: i) collecting the data set might take a long time, and ii) capturing different conditions does not guarantee that the illumination will not change dramatically in test time (i.e., in a manner that will impinge on the inference results). For better generalization, researchers are encouraged to artificially introduce optional illumination conditions along the training stage by using augmentations and regularization methods. However, since the illumination conditions might be too varied, it is important to create a data collection protocol that takes as much of these conditions into account. The augmentation of illumination is not trivial and hence should be carefully used, and only as a supporting process to the training of the model.

**Object occlusions** - This challenge exists in many computer-vision tasks and is amplified in agricultural cases, especially in counting-related applications. The agricultural target object is often occluded by leaves, branches, or other non-target objects (e.g., infrastructure wires in greenhouses). Moreover, the target object might be highly occluded by other target objects, since some fruits (e.g., apples, grapes, and tomatoes) grow in clusters, making it difficult to differentiate between single fruits. Thus, it is mandatory to design the data collection and annotation processes with great care. In choosing a counting methods, the occlusion level of the objects (which is a trait of the environment and the specific cultivar) should be considered. To avoid miss-detection (i.e., missing occluded objects), occluded objects should be annotated as objects, and if the target object grows in a cluster formation, researchers should consider a pre-processing stage of determining the number of objects in a cluster so as to bound the number of possible objects (and prevent double counting).

**Double counting** - Double counting is a common challenge in many counting applications and is particularly relevant in agriculture, where image tiling is considered a common solution for high-resolution images (as mentioned in counting methods section). Tiling the image results in some objects being cut into two or more parts. Thus, when the objects in each tile are counted, the same object might be counted more than once. Therefore, it is essential to re-assemble the image after the counting analysis and then perform an additional Non-Maximum Suppression (NMS) process. Since it is difficult to collect quality datasets, many recent studies have examined the use of automated data collection methods. However, tracking objects in each frame is not an easy task, and it heavily relies on the performance of the object detector (Villacres, 2023). Object tracking becomes even more challenging in real-world environments due to occlusion problems, where objects can be hidden or partially obscured by other objects. This can lead to confusion and mixing up of objects, making it difficult for the tracker to distinguish between them. To overcome these challenges, the choice of the tracker must be carefully considered and tuned to yield the required performance. Although recent work has improved object tracking in real-world environments (Villacres, 2023; Gao 2022), tracking remains a challenging task, particularly in the agricultural domain. This is due to the complexity and variability of agricultural environments, where objects can be highly occluded and exhibit non-rigid motion patterns. As such, further research is needed to develop more robust and reliable object tracking methods for agricultural applications.

**Background vs. foreground** - Analyzing plants and trees in their natural environment gives rise to two very different challenges. The first derives from a very dense scene (e.g., a tree with dense foliage), causing the target object to be highly occluded, whereas the second derives from sparse trees or plants, allowing the background to be visible and, hence, causing the counting algorithm to account for irrelevant objects. This issue is clearly evident in Fig. 4. For dense scenes, a good direction would be to rethink the counting application. For example, if the application is required to count trees, collecting the images from a top view (using a UAV) can better differentiate the trees. For fruits or flowers that grow in close proximity, one to the other, or in cluster formation, combining the detection of clusters with the counting of objects within the clusters can yield good results. For sparse scenes, a naive solution is to use a background screen to ensure that non-relevant objects will not be visible. A better solution is to screen objects beyond a certain distance from the camera (by integrating data from a depth camera sensor with RGB data) so that only objects that are within a specific distance from the camera are counted. This approach poses additional challenges, as we further discuss below.

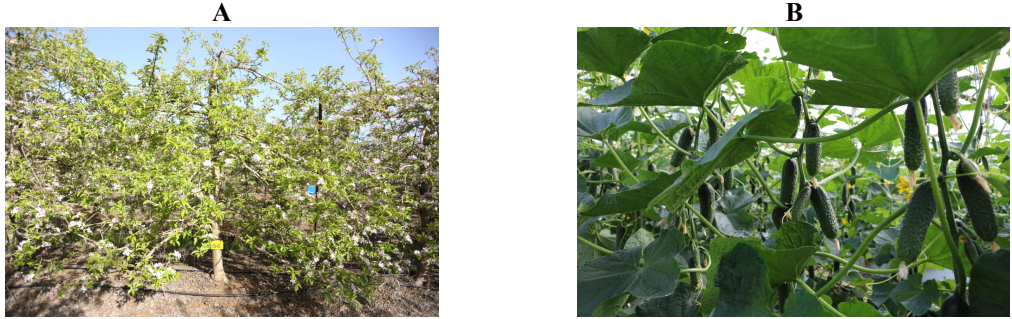

**Fig. 4** The difference between dense and sparse scenes. Fig. 4A shows dense apple flowers in a real outdoor environment. Since the trees are dense, there are many occlusions, and it is difficult to differentiate between the trees. the flowers are highly occluded by the trees' heavy foliage. Fig. 4B shows sparse cucumber plants in a greenhouse. Since the plant is sparse, the row in the background becomes visible, which might cause the model to count the background fruits, the cucumbers in the background are visible, since the plant is relatively sparse.

## Hardware challenges

**High computational burden** - End devices, such as UAVs, IoTs, or ground robots, cannot carry strong computational units, but DL-based counters are computationally heavy and require advanced computational hardware, such as a Graphic Processing Unit (GPU), a Tensor Processing Unit (TPU), and other similar units. This issue is mostly related to the inference processing time. The entire inference pipeline (including the pre-processing, feed-forward, and post-processing of the input) relies on such hardware. A reasonable solution to these conflicting demands is to stream the data to a cloud (or any remote) server that includes such hardware and to analyze the data there. However, this solution introduces an additional challenge related to the communications infrastructure in the field and to the communications hardware on edge devices. For example, a remote ground robot can carry a SIM card and broadcast the data to remote servers, but a stable and high-frequency internet or cellular connection is still mandatory. This might be a good solution if the examined field is located in inhabited areas.

**Registration of sensors** - Most DL networks rely solely on RGB data. However, incorporating additional sensors gives an extra layer of information that derives from these advanced sensors (such as multi-/hyper-spectral, IR, thermal, and others). When using multiple sensors to analyze a single plant, it is essential to register them. Registering the images is the process of aligning the different images so that the same information will be visible in each of them. Since these sensors are not physically connected, a software solution must be used to define where each pixel in image $i$ is located in image $j$. Otherwise, the results of different algorithms cannot be properly aggregated (if the data is analyzed separately and has a post-processing stage of aggregation) or cannot be used at all (in the case that the algorithm simultaneously uses the various images as a single image with multiple channels). This is a challenging task, since the image resolution in each sensor is different (usually thermal and depth sensors, has small resolution than RGB sensors) The solution in the complex agricultural environment is not trivial. Due to extreme weather conditions (e.g., wind), rugged terrain, and movement of plants, methods must be developed for robust sensor registration, since each slight movement causes misalignment of the sensors, resulting in the need to re-calibrate the sensors.

**High cost of accurate sensors** – The high cost of accurate and advanced sensors (Osco, 2021) means that most data in the agricultural sector is still being collected using standard RGB cameras (as can be seen in Table 1). This poses an additional challenge for the use of advanced sensors, since commonly used methods are tailored to RGB cameras and are not applicable for other data types. In addition, benchmarks for advanced sensor data are lacking.

# Concluding Remarks

This article presents a systematic review of research related to counting objects in agriculture in the past decade. Two hundred forty-three papers published from January 2012 to June 2022 were collected and analyzed to obtain an overview of trends as attempts to automate agriculture move from conventional computer-vision algorithms to DL methods and from RGB data alone to RGB data integrated with information from advanced sensors. Different methods for counting objects in agriculture are reviewed, as are annotation procedures and their burden, data collection procedures, relevant performance indicators for various tasks and sub-tasks, and the challenges of incorporating DL-based counters into different applications.

There are many challenges attendant on incorporating DL models into real-world counting applications. Most of the reviewed studies provide results indicating the benefits of moving from conventional computer-vision algorithms to DL-based counters. The review also shows that there is a connection between some of the challenges. For example, the scarcity of advanced and accurate machinery has led to the development of algorithms that use only RGB images and hence can be used only in daylight. The next generation of work should focus on generalizing these methods and accounting for the variability in data by performing R&D that involves various crops in different fields, a variety of growing conditions, and different seasons.

Another important challenge that remains to be addressed is the lack of standardization of methodologies, the implication being that researchers and practitioners are currently required to devote time and resources to reaching a stable system, before they can start solving additional challenges. We therefore recommend that standard benchmarks and datasets be created. Collected datasets should be made public and will enable comparison of the results of newly developed approaches and the building of generalized models. Researchers can also benefit from sharing code, hyper-parameters, and network weights along with complete documentation and tutorials. This approach will provide solid benchmarks and platforms contributing to the development of fast and stable systems that can be deployed in the field for use by farmers.

Finally, as more advanced sensors become standard and widely used, DL-based counters, which benefit from big data, will become more accurate, opening the way to solving complex tasks, such as real-time yield estimation, monitoring of crop growth and plant diseases, stress modeling, and design of selective pesticide treatments. Such complex tasks require massive, well-annotated datasets collected across seasons, growing conditions, and geographical locations.

# Acknowledgements

This research was partially supported by the Phenomics Consortium, Research Innovation Authority Grant, from the Ministry of Science Grant Number 20187 and from Ben-Gurion University of the Negev through the Agricultural, Biological and Cognitive Robotics Initiative, the Marcus Endowment Fund, and the W. Gunther Plaut Chair in Manufacturing Engineering.

# Compliance with Ethical Standards

The authors declare that they have no competing interests. The authors declare that the research did not involve any Human Participants and/or Animals. This research is funded as stated in the acknowledgements section.

# Appendix
## A – Links to public datasets

1. goo.gl/6aS64p

2. https://zenodo.org/record/3361736#.Yny17vNBxhE

3. https://rex.libraries.wsu.edu/esploro/outputs/99900502619401842

4. https://github.com/fu3lab/Scifresh-apple-RGB-images-with-multi-class-label

5. http://data.acfr.usyd.edu.au/ag/treecrops/2016-multifruit/

6. http://www.robotics.unsw.edu.au/srv/dataset/liu2015lightweight.html

7. https://github.com/xixiareone/counting-pigs/tree/master/counting/datasets

8. https://vhasoares.github.io/downloads.html

9. https://onedrive.live.com/?authkey=%21AMtRaqnBYG25eeE&id=5EE9D2EA82EE73C1%212304&cid=

10. https://zenodo.org/record/5092309#.YrBqcBNBxh5EE9D2EA82EE73C1E

11. https://vision.eng.au.dk/oil-radish/

12. https://figshare.com/s/616956f8633c17ceae9b

13. https://zenodo.org/record/3715991#.Yrbc3hNBw1I

14. https://github.com/avadesh02/MangoNet-Semantic-Dataset

15. https://zenodo.org/record/3712808#.YrcDpRNBw1I

16. https://data.mendeley.com/datasets/d7kbzjr83k/1